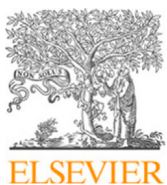
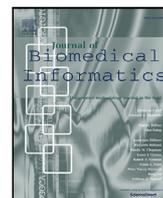

Original Research

# MultiADE: A Multi-domain benchmark for Adverse Drug Event extraction

Xiang Dai [a],[*], Sarvnaz Karimi [a], Abeed Sarker [b], Ben Hachey [c], Cecile Paris [a]

[a] *CSIRO Data61, Sydney, Australia*
[b] *Emory University, Atlanta, United States*
[c] *University of Sydney, Sydney, Australia*

ABSTRACT

**Objective:** Active adverse event surveillance monitors Adverse Drug Events (ADE) from different data sources, such as electronic health records, medical literature, social media and search engine logs. Over the years, many datasets have been created, and shared tasks have been organised to facilitate active adverse event surveillance. However, most – if not all – datasets or shared tasks focus on extracting ADEs from a particular type of text. Domain generalisation – the ability of a machine learning model to perform well on new, unseen domains (text types) – is under-explored. Given the rapid advancements in natural language processing, one unanswered question is how far we are from having a single ADE extraction model that is effective on various *types of text*, such as scientific literature and social media posts.

**Methods:** We contribute to answering this question by building a multi-domain benchmark for adverse drug event extraction, which we named MULTIADE. The new benchmark comprises several existing datasets sampled from different text types and our newly created dataset—CADECv2, which is an extension of CADEC (Karimi et al., 2015), covering online posts regarding more diverse drugs than CADEC. Our new dataset is carefully annotated by human annotators following detailed annotation guidelines.

**Conclusion:** Our benchmark results show that the generalisation of the trained models is far from perfect, making it infeasible to be deployed to process different types of text. In addition, although intermediate transfer learning is a promising approach to utilising existing resources, further investigation is needed on methods of domain adaptation, particularly cost-effective methods to select useful training instances.

The newly created CADECv2 and the scripts for building the benchmark are publicly available at CSIRO's Data Portal (https://data.csiro.au/collection/csiro:62387). These resources enable the research community to further information extraction, leading to more effective active adverse drug event surveillance.

## 1. Introduction

An Adverse Drug Event (ADE) is an injury occurring after the use of a medication. An ADE can be caused by a medication error or a drug interaction (e.g., between multiple drugs) [1,2]. In addition, unexpected harm could be caused by the routine use of medication at the normal dosage; this is termed an Adverse Drug Reaction (ADR) [3]. We refer to these collectively as *adverse event*. Although a dedicated clinical trial phase is required to identify potential adverse drug reactions, not all the adverse effects of a medication are discovered before it goes to market. Such adverse events pose a substantial public health problem, resulting in thousands of incidents of injury, or even death, and considerable financial burdens to healthcare systems [3–5].

There have been a lot of efforts towards the detection, assessment, understanding, and prevention of adverse drug reactions. One key activity is discovering potential adverse effects of medications as early as possible. Spontaneous reporting systems, such as MedWatch,[1] are built to enable health professionals, patients or manufacturers to report side effects they observe or suspect. This approach is called passive surveillance [6]. Additionally, active surveillance has received more attention recently, especially in the medical informatics community, to identify adverse drug events from different sources, such as electronic health records, medical literature, social media and search engine logs [3,4]. One example of these efforts is the release of CADEC – CSIRO

---

* Corresponding author.
*E-mail addresses:* dai.dai@csiro.au (X. Dai), sarvnaz.karimi@csiro.au (S. Karimi).
[1] https://www.accessdata.fda.gov/scripts/medwatch/index.cfm
[2] https://www.askapatient.com/

Adverse Drug Event Corpus – created using data from AskaPatient forum[2] on first-hand experiences from patients on adverse drug events of medications. Alongside CADEC, many new corpora have been released, and shared tasks have been organised to facilitate active surveillance research for adverse drug events [7–10].

One trend we observe from the recent literature on ADE extraction is that most—if not all—datasets or shared tasks focus on extracting adverse drug events from one particular type of text. For example, the 2018 N2C2 shared task on adverse drug events and medication extraction [7] focused on mining adverse drug events from clinical notes; a series of shared tasks organised with the Social Media Mining for Health (SMM4H) workshop [9–14] focus on social media data (e.g., tweets); the TAC 2017 Adverse Reaction Extraction from Drug Labels Track [15] focused on drug labels; and there are also datasets built using scholarly articles [16,17] or online posts from patient forums [18,19]. Domain generalisation – the ability of a machine learning model to perform well on different types of text – is under-explored for ADE extraction.

Following the success of pre-trained models and, more recently, Large Language Models (LLMs), many researchers in the Natural Language Processing (NLP) community have explored how to build a single model capable of solving all NLP tasks on different types of text with minimum effort. The fine-tuning approach, where the pre-trained model is fine-tuned using a small amount of labelled data, has demonstrated state-of-the-art effectiveness on many NLP tasks [20,21]. More recently, the prompt-based approach enables a large language model to solve various tasks by providing task instructions and, optionally, demonstrations [22]. In the information extraction area, Xie et al. [23] explore the reasoning capability of LLM on Named Entity Recognition (NER) and focus on the zero-shot setup using ChatGPT. Han et al. [24] evaluate ChatGPT's performance on 14 information extraction tasks under the zero-shot, few-shot and chain-of-thought scenarios. Their results show that fully supervised NER models (following the fine-tuning approach) still outperform ChatGPT (using prompt engineering) by a large margin. These explorations naturally lead to an important question: how far are we from having a single ADE extraction model that works well on different types of text?

To contribute towards answering this question, we build a multi-domain benchmark for adverse drug event extraction—MULTIADE—and present our experimental results under different evaluation scenarios. The MULTIADE benchmark comprises several existing datasets sampled from various text types—N2C2 [7] and MADE [8] from clinical notes, PHEE [17] from scholarly articles, PSYTAR [19] and CADEC [18] from social media data. In addition, we also created a new dataset: CADECV2, an extension of CADEC, covering online posts regarding more diverse categories of drugs than CADEC. Our benchmark results show that the generalisation of the trained models is far from perfect, making it challenging to deploy them to process text from different sources. Our findings highlight the need for further investigation of domain adaptation methods and transfer learning for adverse drug event extraction.

Our paper is structured as follows: Section 2 reviews recently published ADE extraction datasets and models; Section 3 introduces the newly developed CADECV2 dataset; Section 4 describes the MULTIADE benchmark; Section 5 details our benchmark results; and a summary is provided in Section 6.

| **Statement of significance** | |
|---|---|
| Problem | Can we build a single ADE extraction model that works well on various text types? |
| What is Already Known | Many datasets are created, and shared tasks are organised to facilitate active adverse event surveillance. However, domain generalisation-the ability of a machine learning model to perform well on new, unseen text types-is under-explored. |
| What This Paper Adds | We build a multi-domain benchmark for adverse drug event extraction (MultiADE), comprising several existing datasets sampled from different text types and our newly created dataset-CADECv2. Our benchmark results show that the generalisation of the trained models is far from perfect, making it impossible to be deployed to process different text types. |

## 2. Related work

We survey relevant studies on developing ADE-focused datasets and ADE extraction models. These studies include datasets that are related to ours or propose methods of extracting related entities. We then use these datasets and methods in our benchmarking in the next section.

### 2.1. ADE-focused datasets

We list only recently published (after 2015) English ADE-focused datasets and refer readers to Table 1 in Karimi et al. [18] and Table 1 in Thompson et al. [16] for a review of relevant datasets created before 2015. Although many ADE datasets exist in languages other than English [14,25–27], we believe the challenge of building a single ADE extraction model for different languages is quite different from building a model for different domains. Therefore, we take an incremental step by focusing only on English datasets (see Table 1).

*SMM4H.* The Social Media Mining for Health Applications shared tasks focus on mining health-related information from publicly available user-generated content. ADE mining in tweets is one of the longest-running challenges at the SMM4H shared task [9–14,29]. Three key NLP problems, forming a social media-based pharmacovigilance pipeline, include identifying if a tweet mentions an ADE or not (2016 task1, 2017 task1, 2018 task3, 2019 task1, 2020 task2, 2021 task1 and task2, 2022 task1), extracting the text span of ADEs in tweets (2016 task1, 2019 task2 and task3, 2020 task3, 2021 task1, 2022 task1), and normalising ADE mentions to MedDRA (2017 task3, 2019 task3, 2020 task3, 2021 task1, 2022 task1).

*TAC-2017.* [15] is a dataset used in the Adverse Reaction Extraction from Drug Labels Track as part of the 2017 Text Analysis Conference (TAC). While the ultimate goal is to automate the manual approach to determine if a given adverse event is already noted in the structured product labels, the track evaluates and provides data for several information extraction tasks (e.g., concept identification, concept normalisation and relation extraction). The entity annotations in TAC-2017 include adverse reaction mentions and modifier terms (e.g., negation, severity, and drug class), and relations between them – negated, hypothetical and effect – are annotated. Positive adverse reaction mentions are also normalised to MedDRA, and mappings contain the Lowest Level and Preferred Terms.

*PHAEDRA.* Thompson et al. [16] provide detailed information about the effects of drugs using multiple levels of annotations. Three types of entities – drugs, disorders and medical subjects – are annotated, while drugs and disorders are linked with concept IDs in MeSH and SNOMED-CT, respectively. The PHAEDRA dataset also contains relation annotations that link medical subjects with their conditions (*Subject_Disorder*) and link together different names for the same concept (*is_equivalent*).

**Table 1**
Recently published (after 2015) ADE-focused datasets in English. Note that we only consider labelled data in the size column.

| Corpus | Size | Source |
| --- | --- | --- |
| SMM4H 2016 [11] | 20405 and 2067 tweets | Twitter |
| TAC-2017 [15] | 200 labels | Drug labels |
| SMM4H 2017 [12] | 25678 tweets and 9150 phrases (mentions) | Twitter |
| PHAEDRA [16] | 597 abstracts | MEDLINE |
| SMM4H 2018 [13] | 30633 tweets | Twitter |
| MADE 1.0 [8] | 1089 notes | Discharge summaries, consultation reports, and other clinical notes |
| PsyTAR [19] | 6009 sentences | Online healthcare forum |
| SMM4H 2019 [9] | 30253 and 3252 tweets | Twitter |
| n2c2-2018 [7] | 505 discharge summaries | Discharge summaries |
| SMM4H 2020 [14,14] | 30437 tweets | Twitter |
| ADE Eval [28] | 200 labels | Drug labels (package inserts) |
| PHEE [17] | 4827 sentences | MEDLINE |

*MADE 1.0.* [8] comprises 1089 fully de-identified longitudinal Electronic Health Record (EHR) notes from 21 randomly selected patients with cancer at the University of Massachusetts Memorial Hospital. Nine clinical named entity types (i.e., drugname, dosage, route, duration, frequency, indication, ADE, severity, and other SSD—Sign, Symptom, or Disease) and seven relations (i.e., ADE–drugname, SSD-severity, indication–drugname, drugname–route, drugname–dosage, drugname–duration, drugname–frequency) are annotated in the corpus.

*PsyTAR.* [19,30], as our corpus, is developed using patients' narrative data from a healthcare forum (askapatient.com). However, Zolnoori et al. focus on psychiatric medications, particularly SSRIs (Selective Serotonin Reuptake Inhibitor) and SNRIs (Serotonin Norepinephrine Reuptake Inhibitor). Sentences in the review posts are classified for the presence of adverse drug reactions, withdrawal symptoms, sign/symptom/illness, drug indications, drug effectiveness, and drug ineffectiveness. Four types of concepts – adverse drug reactions, withdrawal symptoms, sign/symptom/illness, and drug indications – are annotated while linked with concept IDs in UMLS and SNOMED-CT. In addition, the entities are further classified as physiological, psychological, cognitive, and functional problems (e.g., limitation in daily functioning, social activities, or interpersonal relationships).

*n2c2-2018.* Track 2 of the 2018 National NLP Clinical Challenges shared task focuses on the extraction of medications and ADEs from clinical narratives [7]. The data for the shared task consists of 505 discharge summaries drawn from the MIMIC-III database [31]. Medication information detailed in the narratives includes medications, their strengths and dosages, duration and frequency of administration, medication form, route of administration, reason for administration, and any observed ADEs associated with each medication.

*ADE eval.* [28] is developed to evaluate NLP techniques for identifying ADEs mentioned in publicly available FDA-approved package inserts (drug labels). The task consists of identifying mentions of ADEs in specific sections of package inserts and mapping those mentions to associated terms in MedDRA terminology.[3] It is worth noting that part of the training data for ADE Eval (50 documents) is from the TAC-2017 test set.

*PHEE.* Sun et al. [17] design a hierarchical schema to provide coarse and fine-grained information about (a) patients, (b) treatments – the therapy administered to the patients, and (c) effects – the outcome of the treatment. Their fine-grained annotations are usually words or short phrases that highlight specific details, such as age, gender, race, number of patients, pre-existing conditions of the patients and drug, dosage, frequency, route, time elapsed, duration, target disorder of the treatment. Note that the PHEE dataset is built on existing corpora (i.e., PHAEDRA [16] and ADE [32]). All sentences in ADE [32] and those in PHAEDRA [16] with *Adverse Effect* and *Potential Therapeutic Effect* annotations are collected and enriched using the proposed annotation schema.

### 2.2. ADE extraction models

Developing automatic methods for detecting and extracting ADEs has a long-standing history [33–35], and methods range from statistics-based [36–38] to machine learning-based [39–41], and more recently deep learning-based [42–48]. Raval et al. [27] frame ADE extraction as a sequence-to-sequence problem using the T5 model architecture [49]. Scaboro et al. [50] compare the effectiveness of different transformer-based models for ADE extraction and use feature importance techniques to correlate model characteristics (e.g., training data, model size) to their effectiveness. Sun et al. [51] investigate the capability of ChatGPT for ADE extraction via various prompts and demonstration selection strategies. Li et al. [52] evaluate several LLMs (e.g., GPT-3, GPT-4 and Llama 2) on extracting influenza vaccine adverse events.

Although there are several ADE extraction models proposed recently, most of these methods are evaluated on a single dataset or datasets sourced from the same text type. Our study builds on previous research [27,50] employing pre-trained transformer models for ADE extraction and takes a step further by focusing on these models' domain generalisation. By benchmarking datasets from different text types, we aim to investigate how feasible it is to build a single ADE extraction model that is effective on various types of text.

### 3. CADECv2

Similar to CADEC [18], we use data from a medical forum—AskaPatient—to build the new corpus. On AskaPatient, the patient can fill out a review form on a specific drug. The form allows both structured and free-text fields. The structured fields include patient information (e.g., age and gender) and dosage details (e.g., quantity, strength, and frequency); the patient can also write the reason for taking the drug, the side effects experienced and comments in free-text format. In CADECv2, we only release and annotate the free text sections of each post.

CADEC [18] covers two types of drugs: (1) those with Diclofenac (an anti-inflammatory drug used to treat pain and inflammatory diseases) in their active ingredients, and (2) Atorvastatin (also known as Lipitor, used to prevent cardiovascular disease). In CADECv2, we obtain data relating to 26 drugs, thus covering more diverse drug categories, including those for depressive disorders (e.g., Duloxetine), hypertension (e.g., Losartan, Valsartan, Valsartan/hydrochlorothiazide, Amlodipine, Valsartan), stomach and esophagus problems (e.g., Esomeprazole, Ranitidine), and Diclofenac.

Table 2 compares statistics on the data used to create CADEC and CADECv2. It is worth noting that most of these reviews in AskaPatient are written by the patient, reporting their personal experience. However, not all patients provided their age and gender. Also, sometimes the conditions of a family member were reported, and not that of the person posting which is why the minimum patient age in CADECv2 is 0. For example, a review starts with "MY 2.5 MONTH OLD BABY TOOK …", indicating an infant.

---
[3] https://www.meddra.org/

Table 2
A comparison of statistics on the data used in CADEC and CADECv2. There are 53 posts containing duplicated content in both datasets. Each post can contain multiple sentences or even paragraphs.

|  | CADEC | CADECv2 |
| --- | --- | --- |
| Number of posts | 1250 | 3548 |
| Average length (word) | 98.8 | 111.3 |
| Number of drugs | 12 (2 groups) | 26 (4 groups) |
| Time span | 2001 Jan–2013 Sep | 2001 Apr–2019 Sep |
| Gender | Female: 662 (50.1%) Male: 617 (49.9%) | Female: 902 (68.5%) Male: 414 (31.5%) |
| Patient age range | 17–84 | 0–91 |

### 3.1. Annotation guidelines

We followed the entity recognition annotation guidelines used to build CADEC and refer readers to Section 4.1 in [18] for a detailed description. We release the complete annotation guidelines together with the annotated corpus. Here, we only provide a brief description of the guidelines for self-containment.

There are, in total, five entity categories we annotated in CADECv2: (1) Drug—mentions of the drug name; (2) Adverse Drug Event—mentions of side effects associated with a drug; (3) Disease—the name of a disease for which the patient takes the drug; (4) Symptom—symptoms of a disease that leads to the patient taking a drug; and, (5) Finding—side effect that is not directly experienced by the patient, or a clinical concept that the annotator is unclear which category it belongs to.

We instructed the annotators to conduct the annotation at the sentence level. That is, no entity mentions should span over sentences. Entity mentions could be discontinuous (i.e., consisting of components separated by intervals) but not nested (i.e., one mention is completely contained by the other). For example, the phrase "have much muscle pain and fatigue" is annotated with two mentions: 'muscle pain' and 'muscle fatigue', where the latter is a discontinuous mention [53].

One major source of disagreement in annotations in CADEC was on deciding the entity mention boundary. For example, when the span matching was strict, the inter-annotator agreement reported in [18] on posts relating to Diclofenac was 46.6. This agreement increased to 68.7, when the span matching is configured to be relaxed—annotations that overlap were counted as a match. We assisted the annotators to decide the mention boundary by adding detailed guidelines. The expectation was that the identified mentions to be easily (excluding unnecessary context) and unambiguously (including necessary context) linked to a unique concept in controlled vocabularies, such as SNOMED Clinical Terms and MedDRA. Therefore, we suggested the annotators remove modifiers once the meaning of identified mentions (the underlying medical concept) is unchanged. For example, 'ruined sex life' preserves all information in 'Completely ruined my sex life'. However, if a modifier provides specific information, it should be kept. For example, all modifiers in 'Occasionally coughing', 'coughing all day', and 'coughing lasts 3 days' provide more specific information than only annotating 'coughing'. In addition, the annotators were instructed to include the severity indicator, such as 'acute' in 'acute stomach pain'.

### 3.2. Annotation process

To set up the annotations, we used a web-based annotation tool—Brat [54]. Three primary annotators with clinical or public health backgrounds and four secondary annotators with computer science backgrounds were involved in the annotation. Each primary annotator annotated more than 1000 posts, and each secondary annotator annotated less than 500 posts.

Before the formal annotation stage, we conducted two rounds of pilot annotations on 100 posts, which were not included in the final corpus. Disagreements were discussed between annotators after each pilot round, and annotation guidelines were adjusted correspondingly. During the formal annotation stage, there are 996 posts annotated by more than one annotator. On these posts, we measured the pairwise inter-annotator agreement using mention-level $F_1$ scores. That is, we considered annotations from one annotator as ground truth and calculated the $F_1$ score of another annotator, whose annotations are treated as system predictions. The macro-averaged inter-annotator agreement (67.4) was higher than the result reported in [18] on the posts relating to Diclofenac (46.6) using the same strict matching setup.

## 4. MULTIADE benchmark

The MULTIADE benchmark, which is proposed in this study, comprises six publicly available datasets for adverse drug event extraction. We standardised the entity types in these datasets in order to use the same entity recognition task for all. We included five existing datasets: N2C2 [7], MADE [8], PHEE [17], PSYTAR [19], CADEC [18], and our newly created dataset, CADECv2.

In addition to the characteristic of being readily accessible to the public, we chose these datasets based on two key considerations: (1) domain diversity and (2) reproducibility. To test domain generalisation and transferability, we strived to cover various text types in MULTIADE: N2C2, and MADE sampled from clinical notes; PHEE from scholarly articles; PSYTAR, CADEC and CADECv2 from online posts, where PSYTAR focuses on psychiatric medications, CADEC covers anti-inflammatory drugs and drugs for cardiovascular disease, and CADECv2 covers more diverse drugs, including those for depressive disorders, acid-related disorders, esophagus problems and hypertension.

To ensure the reproducibility of our benchmark results, we did not choose datasets released through the SMM4H shared tasks because Twitter permits only sharing Tweet IDs in a dataset for others to use. Consequently, subsequent researchers often find it challenging to fully recover the dataset, as deleted tweets by users are irretrievable.

The descriptive statistics of the datasets in MULTIADE are listed in Table 3.

### 4.1. Analysing domain characteristics

To understand the characteristics of the datasets sampled from different text types, we quantified each dataset's vocabulary (words and adverse drug events) richness and measured the similarity across different datasets.

Vocabulary richness is often used to measure the diversity of words used within a given dataset [55]. We randomly shuffled the sentences in each dataset and iterated through all sentences to calculate the number of unique words given the total number of words seen. This procedure was repeated five times, and each dataset's mean was calculated. Similarly, we measured the richness of adverse drug event vocabulary by considering only mentions representing adverse drug events in each dataset. The results are shown in Fig. 1, demonstrating a wider variety of words and adverse drug event mentions in scholarly articles (i.e., PHEE) compared to other text types. Note that clinical notes may have a richer word vocabulary (e.g., N2C2), but their vocabulary to describe adverse drug events is much narrower (i.e., MADE, N2C2).

We also quantified the similarity between different datasets by calculating the vocabulary overlap (words and adverse drug events, respectively) [56]. We built the word vocabulary containing the top 3K most frequent words in each dataset and the adverse drug event vocabulary containing ADE mentions that appeared more than once in each dataset. The results are shown in Fig. 2. We observed datasets sampled from similar text types have substantial vocabulary overlap while they are far more dissimilar to datasets sampled from other text types.

**Table 3**
The descriptive statistics of the datasets in MULTIADE. The numbers in brackets represent the unique count of mentions. spaCy library is used for tokenisation.

|  | N2C2 | MADE | PHEE | PSYTAR | CADEC | CADECV2 |
| --- | --- | --- | --- | --- | --- | --- |
| Text type | clinical notes | clinical notes | scholarly articles | online posts | online posts | online posts |
| # Examples | 505 | 1,089 | 4,827 | 3,147 | 1,250 | 3,548 |
| # Sentences | 66,810 | 61,563 | 4,844 | 3,169 | 7,593 | 22,389 |
| # Tokens | 1,587,566 | 1,216,211 | 107,569 | 60,028 | 123,494 | 395,044 |
| # Entity categories | 9 | 10 | 21 | 4 | 5 | 5 |
| # Mentions (train) | 45,198 | 60,419 | 18,040 | 4,675 | 6,373 | 22,767 |
| (dev) | 5,694 | 7,269 | 6,094 | 687 | 1,313 | 3,242 |
| (test) | 32,881 | 11,295 | 6,166 | 1,352 | 1,420 | 6,582 |
| (unique) | 8,942 | 12,626 | 11,671 | 3,334 | 4,124 | 10,289 |
| # Drugs (train) | 14,317 | 11,985 | 3,596 | – | 1,261 | 3,482 |
| (dev) | 1,907 | 1,516 | 1,211 | – | 251 | 461 |
| (test) | 10,573 | 2,395 | 1,221 | – | 286 | 1,051 |
| (unique) | 2,907 | 1,358 | 1,505 | – | 323 | 628 |
| # ADEs (train) | 839 | 1,352 | 2,802 | 2,924 | 4,427 | 16,442 |
| (dev) | 120 | 157 | 918 | 424 | 898 | 2,340 |
| (test) | 625 | 431 | 908 | 872 | 990 | 4,736 |
| (unique) | 744 | 501 | 3,284 | 2,656 | 3,400 | 8,965 |

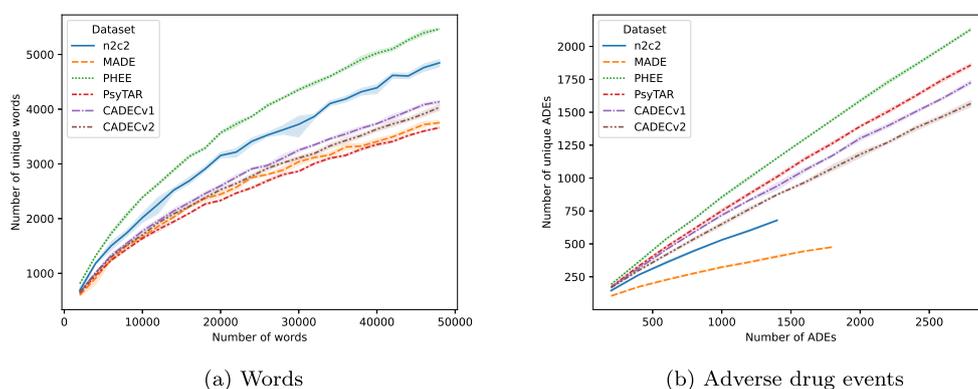

(a) Words  (b) Adverse drug events

**Fig. 1.** A comparison of vocabulary richness between different datasets.

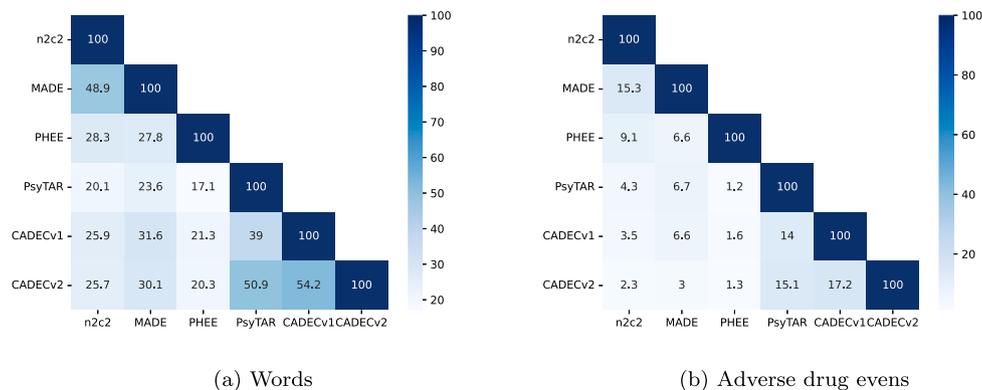

(a) Words  (b) Adverse drug evens

**Fig. 2.** Vocabulary overlap (%) between datasets in MULTIADE. The first two datasets (N2C2 and MADE) are sampled from clinical notes, and the last three datasets (PSYTAR, CADEC, and CADECV2) are sampled from online posts.

*4.2. Standardising the task*

While each dataset is annotated using distinct schemas and guidelines, we aimed to establish a unified benchmark, allowing the evaluation of the model's effectiveness on different text types. Hence, we carefully analysed each dataset's annotation schemas and guidelines, identifying commonalities and differences and linking various entity categories representing closely related concepts (Table 4). For example, the 'Adverse_event' category in PHEE are mainly event triggers, such as 'developed'. This definition is different from the definition of ADE in other datasets. In contrast, 'Effect', indicating the outcome of the treatment, such as 'hemiparesis contralateral to the injury' is closer to the definition of ADE in other datasets. Therefore, we put the 'Adverse_event' category in PHEE as a distinct category (# 15 in Table 4) and link 'Effect' in PHEE and 'ADE' in other datasets (# 2 in Table 4). It is worth noting that most of the datasets also have other types of annotations provided. For example, N2C2 and MADE have relation annotations, e.g., relations between drug names and other mentions such as dosages, duration, etc. CADEC provides concept normalisation annotations that link the identified mentions to controlled vocabularies (i.e., SNOMED Clinical Terms and MedDRA). We leave the investigation of other tasks for future work and focus on identifying mentions of the entities of interest in the MULTIADE benchmark.

Table 4
Annotated entity categories in MULTIADE. Entity categories representing closely related concepts are linked from different datasets. The number in the bracket indicates the number of entity mentions in each dataset. PHI: Protected health information. SSLIF: other signs, symptoms, or diseases. DI: Drug Indications. SSI: Sign/Symptoms/Illness. WD: Withdrawal Symptoms.

|    | N2C2              | MADE              | PHEE                 | PSYTAR     | CADEC         | CADECV2         |
|----|-------------------|-------------------|----------------------|------------|---------------|-----------------|
| 1  | Drug (26800)      | Drug (15902)      | Drug (6028)          | –          | Drug (1800)   | Drug (4995)     |
| 2  | ADE (1584)        | ADE (1940)        | Effect (4630)        | ADR (4220) | ADE (6318)    | ADE (23538)     |
| 3  | Strength (10921)  | –                 | –                    | –          | –             | –               |
| 4  | Form (11010)      | –                 | –                    | –          | –             | –               |
| 5  | Dosage (6902)     | Dose (5694)       | Dosage (460)         | –          | –             | –               |
| 6  | Frequency (10293) | Frequency (4806)  | Freq (114)           | –          | –             | –               |
| 7  | Route (8989)      | Route (2667)      | Route (609)          | –          | –             | –               |
| 8  | Duration (970)    | Duration (898)    | Duration (153)       | –          | –             | –               |
| 9  | Reason (6400)     | –                 | –                    | –          | –             | –               |
| 10 | –                 | Indication (3804) | –                    | DI (769)   | –             | –               |
| 11 | –                 | PHI (84)          | –                    | –          | –             | –               |
| 12 | –                 | SSLIF (39383)     | –                    | –          | Finding (435) | Finding (361)   |
| 13 | –                 | –                 | –                    | SSI (1206) | –             | –               |
| 14 | –                 | Severity (3908)   | –                    | –          | –             | –               |
| 15 | –                 | –                 | Adverse_event (4465) | –          | –             | –               |
| 16 | –                 | –                 | Subject (2394)       | –          | –             | –               |
| 17 | –                 | –                 | Age (690)            | –          | –             | –               |
| 18 | –                 | –                 | Sub-Disorder (365)   | –          | –             | –               |
| 19 | –                 | –                 | Gender (562)         | –          | –             | –               |
| 20 | –                 | –                 | Population (450)     | –          | –             | –               |
| 21 | –                 | –                 | Race (64)            | –          | –             | –               |
| 22 | –                 | –                 | Treatment (5007)     | –          | –             | –               |
| 23 | –                 | –                 | Treat-Disorder (1688)| –          | –             | –               |
| 24 | –                 | –                 | Time_elapsed (312)   | –          | –             | –               |
| 25 | –                 | –                 | Combination (657)    | –          | –             | –               |
| 26 | –                 | –                 | –                    | WD (519)   | –             | –               |
| 27 | –                 | –                 | –                    | –          | Disease (283) | Disease (1577)  |
| 28 | –                 | –                 | –                    | –          | Symptom (275) | Symptom (2142)  |

## 5. Experimental results

The new MULTIADE benchmark enabled us to conduct experiments and analyses regarding domain generalisation and transferability. More specifically, we answered two research questions: (1) how an entity recognition model, which is trained on one text type, generalises to other text types (cross-domain generalisation); and, (2) which text type is the most beneficial transfer source, even though the types of annotated entities may differ between the source and the target (intermediate transfer learning).

Here, we first provide an overview of our entity recognition models. We mainly considered one widely used entity recognition model – span-based entity recognition model [57] – because of its effectiveness in recognising both flat and nested entity mentions, which exist in most of the datasets in the MULTIADE benchmark. We also experimented with a generative entity recognition model [58]. However, its effectiveness was overall lower than the span-based entity recognition model, especially when the input text was long (i.e., N2C2, MADE). We refer the reader to the original work for more details and provide only condensed summaries. We also reported the effectiveness of off-the-shelf LLMs (i.e., GPT-4 [59] and Llama-3-70B [60]) through prompting for the reference. That is, we followed the few-shot setting in [61], where the input prompt consists of task instruction, 5 training examples and the test example, and the LLM was prompted to generate results in the JSON format.

*Span-based entity recognition model.* The key idea of the span-based entity recognition model is to enumerate all possible spans (i.e., continuous text segments) and determine whether each span is a valid entity name and its entity category [62,63]. We first employed a transformer-based encoder, RoBERTa [21], to obtain the contextual vector representations for each token in the text. Then, the vectors corresponding to two boundary tokens of the span and a dense representation of the span length were concatenated and taken as the input of a classifier. The classifier finally predicted either (1) an entity category $t \in \mathcal{R}$; or, (2) NA (the span is not an entity mention).

*Generative entity recognition model.* Instead of a classification problem, the generative model formulates the entity recognition task as a sequence generation problem, which can be solved using a sequence-to-sequence framework. Following [58], we used the BART [64] model as the backbone model and generate the target sequence in an autoregressive manner. The model takes a sequence of text as input and generates a list of entities. These are represented using the position index of two boundary tokens of the entity and the entity category. For example, the output '0 3 Drug' indicates that a drug name is mentioned in the input text; the first token is its starting token, and the fourth is its ending token. If no entity is mentioned in the input text, the prediction is an empty sequence containing two special symbols: <s>, indicating the start of the sequence and </s>, the end.

### 5.1. Cross-domain generalisation

In the set of experiments regarding cross-domain generalisation, we fine-tuned all models using the training set from one source and evaluate the model on the test set in each target. The source development set was used for hyper-parameter tuning (i.e., learning rate and the number of training epochs) and choosing the best model checkpoint (Fig. 3(a)). It is worth noting that it is impractical to expect a model trained to identify a particular entity category to recognise mentions belonging to other categories. Therefore, we focused on two entity categories (drug and adverse drug events), as these were the most common categories annotated in the different datasets. Models were trained separately to recognise drug names and adverse drug events. Evaluation results on recognising drug names and adverse drug events can be found in Tables 5 and 6, respectively.

The first observation from the experimental results is that there is a clear effectiveness drop when applying models trained on one text type to other text types. For example, the span-based drug name recognition model (Table 5), trained on N2C2, achieved a very high $F_1$ score (92.2%) on the N2C2 test set. However, its effectiveness dropped dramatically when applied to other text types (e.g., $F_1$ score of 83.1% on PHEE). This effectiveness drop was more evident in the adverse drug event recognition models. For example, the adverse drug event recognition

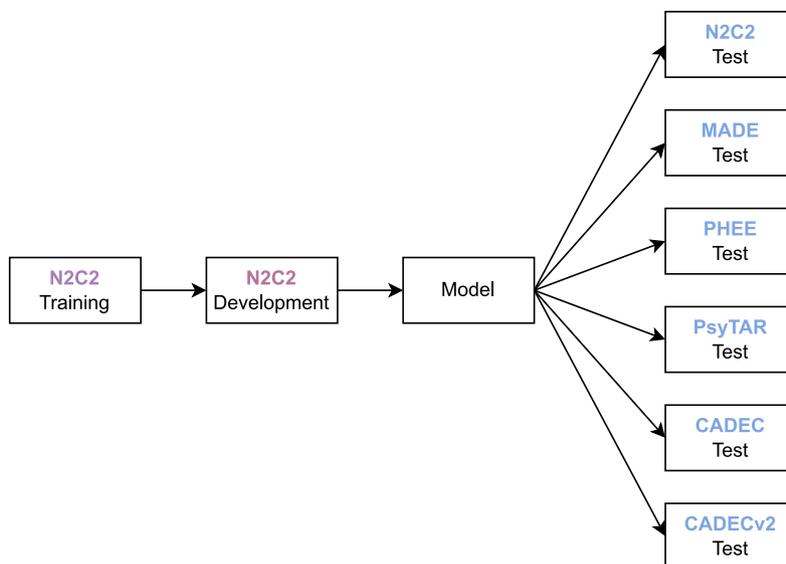

(a) We train the model on one source (e.g., N2C2) and evaluate it on different targets. We aim to analyse how a model trained on one text type generalises to text types.

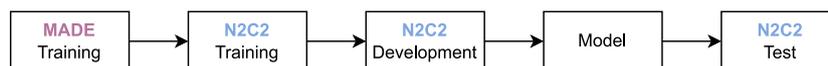

(b) We first train the model on one source (e.g., MADE), then continue training on the target training set (e.g., N2C2), and finally evaluate it on the target test set. We aim to identify the most beneficial transfer source.

**Fig. 3.** The experimental setup of (a) cross-domain generalisation and (b) intermediate transfer learning. The training set is used to train the model, and the development set is used for hyper-parameter tuning and choosing the best checkpoint. Finally, the chosen checkpoint is evaluated on the test set.

**Table 5**
Evaluation results, in terms of $F_1$ scores (in percentage), on recognising Drug names. The PsyTAR dataset does not have 'Drug' annotations.

| Training | Test | | | | | | |
|---|---|---|---|---|---|---|---|
| | N2C2 | MADE | PHEE | PsyTAR | CADEC | CADECv2 | AVG |
| | LLM with in-context learning (5-shot) | | | | | | |
| GPT-4 | 83.9 | 77.6 | 85.3 | – | 81 | 83.3 | 82.2 |
| Llama-3 | 86.1 | 82 | 83.1 | – | 71.7 | 75.6 | 79.7 |
| | Fine-tuned span-based models | | | | | | |
| N2C2 | **92.2 ± 0.1** | 84.5 ± 1.1 | 83.1 ± 0.3 | – | 88.0 ± 0.6 | 84.8 ± 0.8 | **86.5** |
| MADE | 85.6 ± 0.5 | **91.8 ± 0.2** | 79.9 ± 2.4 | – | 88.4 ± 0.6 | 85.2 ± 0.5 | 86.2 |
| PHEE | 75.7 ± 0.9 | 69.6 ± 0.5 | **89.2 ± 0.5** | – | 83.8 ± 0.3 | 80.9 ± 0.2 | 79.8 |
| CADEC | 71.9 ± 1.4 | 65.8 ± 1.0 | 70.9 ± 4.0 | – | **96.2 ± 0.2** | 89.0 ± 0.3 | 78.8 |
| CADECv2 | 73.2 ± 1.5 | 67.0 ± 0.6 | 76.0 ± 0.9 | – | 92.0 ± 0.3 | **91.7 ± 0.0** | 80.0 |
| | Fine-tuned generative models | | | | | | |
| N2C2 | 30.8 ± 43.5 | 29.0 ± 41.0 | 27.6 ± 39.0 | – | 29.1 ± 41.2 | 28.7 ± 40.4 | 29.0 |
| MADE | 57.2 ± 40.4 | 60.9 ± 43.1 | 52.1 ± 36.9 | – | 56.8 ± 40.2 | 55.5 ± 39.2 | 56.5 |
| PHEE | 48.5 ± 10.1 | 42.3 ± 11.9 | 88.9 ± 1.2 | – | 62.6 ± 15.9 | 59.7 ± 15.9 | 60.4 |
| CADECv1 | 64.8 ± 4.9 | 63.7 ± 3.6 | 79.8 ± 0.9 | – | **96.2 ± 0.6** | 89.8 ± 0.5 | 78.9 |
| CADECv2 | 42.5 ± 30.3 | 40.4 ± 28.9 | 50.8 ± 36.0 | – | 60.7 ± 43.0 | 60.3 ± 42.7 | 51.0 |

model trained on N2C2 achieved an $F_1$ score of lower than 10 on other text types (PHEE, PsyTAR, CADEC, CADECv2).

We also observed that models trained on one text type perform better on datasets sampled from a similar text type than a distinct text type. For example, the adverse drug event recognition model, trained on CADECv2, performed reasonably well on PsyTAR, because both CADECv2 and PsyTAR were sampled from online posts even though they focus on different medications. This result is encouraging because the adverse drug events overlap between these datasets is very low (Fig. 2). However, the model can still recognise unseen adverse drug events based on the context in which these events are mentioned. In contrast, its effectiveness degraded when the model was applied to clinical notes (N2C2, MADE) and scholarly articles (PHEE).

Finally, we observed that both GPT-4 and Llama-3 achieve competitive effectiveness in recognising drug names by using only five training examples. On average, GPT-4 achieved the highest $F_1$ score (51.7) on recognising adverse drug events, outperforming the second-best model, which was fine-tuned on CADECv2 with a large margin. However, it is worth noting that fine-tuned models still achieved the best result on each individual dataset.

These results echo the challenge of building a single model that works effectively on different text types, including scholarly articles, clinical notes and online posts [55,65]. Therefore, we move to the next

X. Dai et al.                                                                                           Journal of Biomedical Informatics 160 (2024) 104744Table 6
Evaluation results, in terms of $F_1$ scores (in percentage), on recognising Adverse drug events.

| Training | Test | | | | | | |
|---|---|---|---|---|---|---|---|
| | N2C2 | MADE | PHEE | PSYTAR | CADEC | CADECV2 | AVG |
| | LLM with in-context learning (5-shot) | | | | | | |
| GPT-4 | 25.4 | 64.4 | 38.7 | 58.8 | 62.1 | 61 | **51.7** |
| Llama-3 | 2.8 | 3.5 | 9.1 | 57.7 | 57.5 | 62.4 | 32.2 |
| | Fine-tuned span-based models | | | | | | |
| N2C2 | **51.8 ± 1.2** | 50.3 ± 5.7 | 34.5 ± 6.3 | 0.2 ± 0.2 | 5.2 ± 2.4 | 3.9 ± 1.2 | 24.3 |
| MADE | 38.5 ± 0.8 | **68.7 ± 1.8** | 23.9 ± 4.3 | 1.5 ± 0.2 | 11.7 ± 0.5 | 12.0 ± 0.1 | 27.5 |
| PHEE | 7.7 ± 0.4 | 10.5 ± 0.2 | 69.0 ± 0.5 | 6.5 ± 1.6 | 13.1 ± 1.3 | 10.3 ± 1.1 | 19.5 |
| PSYTAR | 4.3 ± 0.4 | 5.5 ± 0.7 | 27.6 ± 3.3 | **68.5 ± 0.4** | 59.0 ± 0.6 | 58.3 ± 0.7 | 37.2 |
| CADEC | 5.9 ± 0.3 | 11.7 ± 0.5 | 22.5 ± 4.2 | 54.2 ± 1.3 | 67.7 ± 0.6 | 68.9 ± 0.2 | 38.5 |
| CADECV2 | 5.6 ± 0.6 | 13.6 ± 1.7 | 38.7 ± 2.6 | 57.8 ± 0.3 | **69.4 ± 0.2** | **72.0 ± 0.8** | 42.9 |
| | Fine-tuned generative models | | | | | | |
| N2C2 | 15.3 ± 21.6 | 14.1 ± 19.9 | 13.5 ± 19.0 | 0.8 ± 1.1 | 1.1 ± 1.6 | 1.6 ± 2.3 | 7.7 |
| MADE | 20.0 ± 14.2 | 41.1 ± 29.1 | 21.7 ± 15.3 | 4.5 ± 3.2 | 5.8 ± 4.2 | 8.8 ± 6.2 | 17.0 |
| PHEE | 4.4 ± 0.7 | 9.0 ± 2.1 | **71.3 ± 0.4** | 11.0 ± 0.7 | 14.3 ± 0.7 | 11.5 ± 0.1 | 20.3 |
| PSYTAR | 2.4 ± 0.1 | 4.7 ± 0.3 | 42.0 ± 0.1 | 68.2 ± 1.1 | 58.0 ± 0.6 | 55.1 ± 0.2 | 38.4 |
| CADECV1 | 4.4 ± 0.4 | 11.3 ± 1.3 | 37.4 ± 2.9 | 56.6 ± 0.2 | 67.5 ± 0.2 | 66.6 ± 0.2 | 40.6 |
| CADECV2 | 3.6 ± 0.1 | 10.6 ± 0.5 | 37.2 ± 0.9 | 57.7 ± 0.3 | 67.2 ± 0.1 | 69.8 ± 0.3 | 41.0 |

Table 7
Intermediate transfer learning using the span-based model: fine-tuning the model first on the source (row) training data and then on the target (column) training data. The best model checkpoint is chosen based on the results of the target development set and is finally evaluated on the target test set. ↑: better than the baseline model, which is only trained on the target training data. All the numbers are Micro-F1 scores in percentage with standard deviations.

| | N2C2 | MADE | PHEE | PSYTAR | CADEC | CADECV2 | AVG |
|---|---|---|---|---|---|---|---|
| Baseline | 88.1 ± 0.1 | 85.2 ± 0.2 | 67.5 ± 0.0 | 62.1 ± 0.4 | 71.1 ± 0.6 | 72.9 ± 0.3 | 74.5 |
| Combined | 84.5 ± 0.4 | 82.6 ± 0.1 | 66.6 ± 0.3 | 48.2 ± 1.5 | 71.9 ±0.3 ↑ | 70.6 ± 0.1 | 70.7 |
| N2C2 | 88.2 ±0.2 ↑ | **85.6 ± 0.1** ↑ | 67.4 ± 0.4 | 63.4 ±1.0 ↑ | 70.5 ± 0.4 | 72.4 ± 0.3 | 74.6 |
| MADE | **88.4 ± 0.0** ↑ | 85.4 ±0.3 ↑ | 67.4 ± 0.2 | 63.1 ±0.2 ↑ | 71.2 ±0.8 ↑ | 72.7 ± 0.2 | 74.7 |
| PHEE | 88.1 ± 0.1 | 85.3 ±0.1 ↑ | **67.7 ± 0.0** ↑ | 62.3 ±0.3 ↑ | 71.5 ±0.8 ↑ | 72.8 ± 0.1 | 74.6 |
| PSYTAR | 88.3 ±0.0 ↑ | 84.9 ± 0.1 | 67.6 ±0.1 ↑ | 62.1 ± 0.5 | 70.9 ± 0.2 | 72.6 ± 0.4 | 74.4 |
| CADEC | **88.4 ± 0.1** ↑ | 85.3 ±0.2 ↑ | 67.4 ± 0.1 | **63.9 ± 0.7** ↑ | 71.0 ± 0.4 | **73.3 ± 0.2** ↑ | 74.9 |
| CADECV2 | 88.2 ±0.1 ↑ | 85.1 ± 0.3 | 67.4 ± 0.1 | 63.4 ±1.0 ↑ | **72.9 ± 0.4** ↑ | 72.8 ± 0.1 | **75.0** |

question: if we aim to build a model for a particular text type, can we benefit from annotated data from other text types? If yes, which text type is the most beneficial transfer source?

### 5.2. Intermediate transfer learning

In the intermediate transfer learning experiments, we first trained the model on one source training set, then continued training on the target training set, and finally evaluated the model on the target test set. Previous studies [66,67] have observed improvements from intermediate transfer learning, even though the source and the target may have different entity categories. In these experiments, we trained on the source training set using the same set of hyper-parameters, then tuned the hyper-parameters and choose the best model checkpoint based on the results on the target development set (Fig. 3(b)).

We used the model that was only trained on the target training data as the baseline; because if the source and the target are the same, it is equivalent to training on the target training set for longer. Results in Table 7 show that training longer does not change the evaluation results too much compared to the baselines because we have individually conducted hyper-parameter tuning on each baseline model.

We also trained one joint model using all training data from different datasets. Results ('Combined' row in Table 7) show that the model, although trained on much larger training data, underperforms these baseline models with a large gap on most of these target sets. In contrast, we observed overall improvements due to intermediate transfer learning. On five of six datasets, training on a different source outperforms training on only target training data. For example, if a model is first trained on N2C2 and then MADE, it achieves a higher $F_1$ score than training only on MADE. Again, we observed datasets from similar text types can be more helpful than datasets from dissimilar text types. For example, N2C2 and MADE (clinical notes) benefited more from each other than other sources. Similarly, the most beneficial source for CADEC was CADECV2, and vice versa.

These results echoed the necessity of a cost-effective way to select training data [68,69] because much larger training data ('Combined' row in Table 7) did not necessarily outperform the model trained on a small amount of training data ('Baseline'), and training data from close-related domains improved the effectiveness via providing more diverse examples.

### 5.3. Qualitative analysis

To understand the limited generalisation of supervised ADE models, we collected and analysed error predictions on both PSYTAR and PHEE test sets. Based on the vocabulary overlap shown in Fig. 2, these two datasets are the most dissimilar pair, with one built from scholarly articles and the other from online posts. The span-based ADE extraction model trained on PHEE training set achieved an averaged $F_1$ score[4] of 69.0% on the PHEE test set and 6.5% on the PSYTAR test set. Conversely, the model trained on PSYTAR training set achieved an averaged $F_1$ score of 27.6% on the PHEE test set and 68.5% on the PSYTAR test set. First, we analysed common ADE error predictions by all six trained models – three trained on PHEE training set and three on PSYTAR training set using different random seeds – on each test set. On both PHEE and PSYTAR test sets, we observed a small number of false positives but many false negatives, resulting in a low recall.

---

[4] We repeated the experiments three times using different random seeds.

We found that most of these false positives involve minor boundary problems. For example, a model prediction "increase in risk of bleeding" versus the human annotation of "an increase in risk of bleeding", or a model prediction of "completely premorbid" versus the human annotation of "premorbid" led to a mismatch. Severity indicators also caused some error predictions. Although an entity category, 'Severity_cue', is defined in the PHEE annotation schema, they are sometimes annotated as part of ADEs, such as 'severe rhabdomyolysis water intoxication', but not always. On PHEE, we found models perform worse in recognising the potential therapeutic effects (i.e., a potential beneficial effect), which are annotated together with adverse drug effects under the same category 'Effect'. On PsyTAR, the main challenge comes from recognising ADEs written in informal language, such as 'want to express myself and cry but can't', 'alter your thinking'. Note that PsyTAR focuses on psychiatric medications; this may add additional challenges as some ADEs may be very similar to people's daily language and thus only noticeable given a specific context (e.g., 'speeding' in the sentence 'Never have I had such side effects, nor have I ever felt as "drugged" as I did on Cymbalta: nervous and speeding.').

Secondly, we analysed error predictions by out-of-domain trained models but not in-domain models. For example, on the PsyTAR test set, we focused on errors made by models trained on PHEE, but not by models trained on PsyTAR. The model trained on PHEE had a large effectiveness drop when evaluated on the PsyTAR test set. In addition to the challenge of recognising ADEs written in informal language, the other reason for this drop could be the disparity between the sufficient context in scholarly articles and the insufficient context in online posts. We conjecture that models trained on PHEE may learn to rely on context clues (e.g., 'associated with', 'developed') to recognise ADEs because, in scholarly articles, these clues are frequently used to indicate the relation between drugs/treatment and ADEs. However, these clues may not be available in user-generated posts. For example, in PsyTAR, many examples contain only a list of ADEs without providing any context. Conversely, models trained on PsyTAR may not be able to use context clues and fail to recognise unseen ADEs written using a very formal vocabulary. In addition, we noticed that many error predictions are caused by different annotation rules involving coordination structures. That is, PsyTAR usually annotates each ADE separately, but a coordination phrase mentioning several ADEs is usually annotated as one entity in PHEE. For example, in the sentence from PsyTAR "Although if it wasn't for the hair loss and sexual side effects I would of stayed on it.", there are two human annotations: 'hair loss' and 'sexual side effects'. However, models trained on PHEE tend to recognise "hair loss and sexual side effects" as one ADE.

## 6. Conclusions

We built a multi-domain benchmark for adverse drug event extraction—MultiADE. The new benchmark comprises several existing datasets sampled from different text types and a newly created dataset—CADECv2, an extension of CADEC [18], covering online posts regarding more categories of drugs.

Our analysis showed that datasets sampled from different text types have notable differences regarding their vocabulary richness, and datasets sampled from similar text types have substantial vocabulary overlap. Our benchmark results also indicated that the generalisation of the trained model is far from perfect, making it infeasible to be deployed to process different text types. LLMs (e.g., GPT-4) showed more robust results across different datasets; however, their effectiveness in recognising adverse drug events is still far from perfect, underperforming domain-specific supervised fine-tuned models, especially when a large amount of in-domain training data are available.

Although intermediate transfer learning is a promising approach to utilising existing resources, further investigation on methods of domain adaptation, particularly cost-effective methods to select useful training instances, is needed. Another avenue to extend our study is to build similar benchmarks focusing on relation extraction and concept normalisation. However, new datasets need to be annotated with relations between entities and associate the current entity mentions to their corresponding concepts in ontologies such as SNOMED CT or MedDRA.

## Limitations

To investigate domain generalisation, we build a multi-domain benchmark for adverse drug event extraction. Except for one newly created dataset (i.e., CADECv2), this benchmark reuses several existing datasets that are annotated using different annotation schemas. Although we carefully analyse each dataset's annotation schemas and guidelines and try to link categories representing closely related concepts, we know this approach can be imperfect. We call for research to consider reusing existing annotation guidelines or building multiple domain datasets using the same annotation guideline.

## CRediT authorship contribution statement

**Xiang Dai:** Writing – original draft, Visualization, Formal analysis, Data curation, Conceptualization. **Sarvnaz Karimi:** Conceptualization. **Abeed Sarker:** Data curation. **Ben Hachey:** Writing – review & editing. **Cecile Paris:** Conceptualization.

## Ethics

This work is approved by the CSIRO Health and Medical Human Research Ethics Committee (2020_004_LR).

## Declaration of competing interest

The authors declare that they have no known competing financial interests or personal relationships that could have appeared to influence the work reported in this paper.

## Acknowledgements

The authors thank Shanae Burns (CSIRO), Annette Quansah and Zhanxu Liu (Emory University) for their contributions to annotation and discussions and AskaPatient.com for providing their data for use in this study. This work was funded by the CSIRO Precision Health Future Science Platform.